\documentclass[runningheads]{llncs}
\usepackage[T1]{fontenc}
\usepackage{graphicx}
\usepackage{enumitem}
\usepackage{graphicx}
\usepackage{caption}
\usepackage{subcaption}
\usepackage[margin=1in]{geometry}
\usepackage{amsmath,amsfonts,amssymb}
\usepackage{graphicx}
\usepackage[colorlinks=true, allcolors=blue]{hyperref}
\usepackage{booktabs}
\usepackage{subcaption} 
\usepackage{multirow}
\usepackage{xcolor}
\begin{document}
\title{Adaptive Dataset Quantization: A New Direction for Dataset Pruning}
%
%
\author{YU CHENYUE\inst{1} \and
Yu Jianyu\inst{2} }
\authorrunning{YU CHENYUE and Yu Jianyu}
%
\institute{National University of Singapore, Singapore 
\email{e1143627@u.nus.edu} 
\and
PetroChina Company Limited, Cangzhou, China 
\email{kqfw-yjy@petrochina.com.cn}}

\maketitle              
\begin{abstract}
This paper addresses the challenges of storage and communication costs for large-scale datasets in resource-constrained edge devices by proposing a novel dataset quantization approach to reduce intra-sample redundancy. Unlike traditional dataset pruning and distillation methods that focus on inter-sample redundancy, the proposed method compresses each image by reducing redundant or less informative content within samples while preserving essential features. It first applies linear symmetric quantization to obtain an initial quantization range and scale for each sample. Then, an adaptive quantization allocation algorithm is introduced to distribute different quantization ratios for samples with varying precision requirements, maintaining a constant total compression ratio. The main contributions include being the first to use limited bits to represent datasets for storage reduction, introducing a dataset-level quantization algorithm with adaptive ratio allocation, and validating the method's effectiveness through extensive experiments on CIFAR-10, CIFAR-100, and ImageNet-1K. Results show that the method maintains model training performance while achieving significant dataset compression, outperforming traditional quantization and dataset pruning baselines under the same compression ratios.

\end{abstract}
\section{INTRODUCTION}
\label{sec:intro}  
The rapid growth of datasets has driven the success of deep neural networks (DNNs) across various domains~~\cite{deng2009imagenet}. However, large-scale datasets demand significant storage, posing challenges for edge devices with limited capacity~~\cite{jouini2024survey}. Image datasets collected from remote monitoring, drone surveillance, and industrial cameras also face bandwidth constraints, with scale further increasing communication costs.

To address these storage and communication issues, a number of \textit{dataset condensation} methods have been proposed. Notably, \textbf{dataset pruning}~\cite{yang2022dataset} reduces storage by selecting a subset of informative samples, while \textbf{dataset distillation}~\cite{lei2023comprehensive} synthesizes compact representative samples that preserve task-specific performance. Both approaches aim to reduce dataset size by eliminating \textit{inter-sample redundancy}, i.e., minimizing redundancy between images.

However, these methods typically treat each image as an indivisible unit, ignoring the fact that \textit{intra-sample redundancy} also contributes significantly to dataset size. Redundant or less informative content may exist within each image—such as repetitive textures, background pixels, color precision, or high-frequency noise—that may be unnecessary for model training.

Therefore, reducing the storage of each image itself, while preserving essential semantic and structural features, provides a promising complementary strategy to further compress datasets and enhance training efficiency, particularly in resource-constrained deployment settings.

Quantization reduces data from high precision (e.g., float32) to low precision (e.g., int8) to save storage. While model quantization has been widely adopted to compress deep learning models~\cite{rokh2023comprehensive}, unlike model quantization, which benefits from a trainable pipeline with gradient-based feedback and adaptive learning, dataset quantization is a preprocessing-only step with no feedback loop or trainable parameters, making it inherently more challenging~\cite{li2024contemporary}.  

In this paper, we propose a novel algorithm—dataset quantization—that compresses the training set by reducing intra-sample redundancy, while preserving essential features and maintaining training performance. Specifically: 1) We first apply linear symmetric quantization to each sample to obtain an initial quantization range and scale. 2) To accommodate varying precision requirements across samples, we propose an adaptive quantization allocation algorithm that distributes different quantization ratios while maintaining a constant total compression ratio.

Our main contributions are summarized as follows: 1) To our knowledge, this is the first work to propose a solution using a limited bit to represent datasets, aiming to reduce storage requirements. 2) We introduce a dataset-level quantization algorithm with adaptive quantization ratio allocation to get a better result compared with traditional quantization algorithms. 3) Extensive experiments on various datasets including CIFAR-10, CIFAR-100 and ImageNet-1K have validated the effectiveness of our method.

\section{RELATED WORK}
\subsection{DATASET PRUNING}
Dataset pruning, or coreset selection, reduces dataset storage by selecting informative samples based on predefined criteria. GraNd/EL2N~\cite{paul2021deep} ranks samples by gradient magnitude, while TDDS~\cite{zhang2024spanning} uses a dual-depth strategy to capture both training dynamics and sample representativeness. CCS~\cite{zheng2022coverage} employs stratified sampling to improve performance under high pruning ratios. Entropy~\cite{coleman2019selection} selects high-uncertainty samples, and Forgetting~\cite{toneva2018empirical} tracks misclassification events during training. AUM~\cite{pleiss2020identifying} detects mislabeled or ambiguous data based on prediction margins. $D^2$~\cite{maharana2023d2} uses forward and
 reverse message passing over this dataset graph for coreset selection. However, these methods focus solely on reducing the number of samples, overlooking the potential to compress each sample to further reduce dataset storage.
\subsection{DATASET DISTILLATION} 
Dataset distillation synthesizes a compact proxy dataset—often initialized from noise—that preserves the distributional characteristics and empirical training performance of a much larger original set. For example, G-VBSM~\cite{shao2024generalized} aligns synthetic samples with full-data score functions across architectures; SRe2L ~\cite{shao2024generalized} decouples bilevel optimization by squeezing prototypes, recovering richer representations, and relabeling them; MTT~\cite{cazenavette2022dataset} optimizes distilled data so models trained on it follow parameter trajectories similar to those induced by real data; and DM~\cite{zhao2023dataset} directly matches feature-space statistics (means, covariances) between synthetic and real images. By reducing the number of samples rather than compressing each one, these approaches achieve substantial dataset-size reduction without sacrificing accuracy.
\subsection{QUANTIZATION IN DEEP LEARNING} 
Quantization is a widely used technique in deep learning compression that converts high-precision floating-point values into low-precision integers to reduce storage and accelerate inference. LLSQF~\cite{zhao2020linear} proposes a learned linear symmetric quantizer for neural network processors, enabling low-bit quantization of weights and activations. LG-LSQ~\cite{lin2022lg} introduces a scaling simulated gradient (SSG) method to optimize the quantization scaling factor during training. Random Projection~\cite{li2019random} provides a comprehensive analysis of the bias and variance in various quantized estimators. VLCQ~\cite{abdel2025vlcq} extends beyond fixed-point encoding by adopting variable-length coding, allowing more flexible quantization schemes.
While these methods focus on model compression, to the best of our knowledge, no existing work has explored quantization as a means of reducing dataset storage while maintaining model performance.
  
\section{Problem Formulation}
Throughout this paper, we denote the full training dataset as \( \mathcal{D} = \{(d_n, y_n)\}_{n=1}^N \), where \( d_n \in \mathbb{R}^{H \times W \times C} \) is an image sample and \( y_n \in \{1,\dots,C\} \) is its label. Let \( f(\cdot; \theta) \) be a neural network with parameter \( \theta \), and \( \mathcal{L}(\mathcal{D}; \theta) \) be the empirical risk over dataset \( \mathcal{D} \).

To reduce dataset storage, we construct a quantized dataset \( \tilde{\mathcal{D}} = \{(\tilde{d}_n, y_n)\}_{n=1}^N \), where \( \tilde{d}_n \) is the dequantized reconstruction from quantized representation \( d_q \) using a sample-specific bit-width \( b \). Our objective is to ensure that training on \( \tilde{\mathcal{D}} \) yields comparable generalization performance to the original dataset \( \mathcal{D} \):

\begin{equation}
\mathbb{E}_{(x,y)\sim \mathcal{D},\ \theta_0 \sim \mathcal{P}_{\theta_0}} \left[ \mathcal{L}\left(f(\tilde{d}; \theta_{\tilde{\mathcal{D}}}),\ y \right) \right] 
\simeq 
\mathbb{E}_{(x,y)\sim \mathcal{D},\ \theta_0 \sim \mathcal{P}_{\theta_0}} \left[ \mathcal{L}\left(f(d; \theta_{\mathcal{D}}),\ y \right) \right]
\end{equation}

where \( \theta_{\mathcal{D}} \) and \( \theta_{\tilde{\mathcal{D}}} \) denote the final model parameters obtained by training on \( \mathcal{D} \) and \( \tilde{\mathcal{D}} \), respectively, both initialized from \( \theta_0 \sim \mathcal{P}_{\theta_0} \).

The goal of dataset quantization is thus to minimize the storage cost while preserving training effectiveness:
\begin{equation}
\min_{\tilde{\mathcal{D}}} \quad \text{Storage}(\tilde{\mathcal{D}}) \quad \text{s.t.} \quad 
\text{Acc}(f(\cdot; \theta_{\tilde{\mathcal{D}}})) \simeq \text{Acc}(f(\cdot; \theta_{\mathcal{D}}))
\end{equation}
 
\section{METHODOLOGY}
\label{sec:method}

\subsection{Linear Symmetric Quantization}

To compress the dataset and reduce memory footprint, we adopt \textbf{linear symmetric quantization}~\cite{zhao2020linear}, a widely used uniform quantization method. For each image sample \( d \in \mathbb{R}^{H \times W \times 3} \) in the dataset \( \mathcal{D} \), we flatten it into a 1D vector \( \mathbf{z} = [z_1, z_2, \ldots, z_n] \in \mathbb{R}^n \), where \( n = H \times W \times C \). We first compute the maximum absolute value:
\begin{equation}
    m_d = \max_i |z_i|,
\end{equation}
where \( m_d \) captures the dynamic range of the sample. We then define a sample-specific scale factor \( s_d \) as:
\begin{equation}
    s_d = \frac{m_d + \epsilon}{2^{b-1} - 1}, \quad \epsilon > 0 \ (\text{e.g., } 10^{-12}),
\end{equation}
where \( b \) is the quantization bit-width and \( \epsilon \) prevents division by zero. This ensures that all elements can be linearly mapped to the target quantization range.

Each element \( z_i \) is then quantized to a signed integer \( z_{q,i} \in [-Q, Q] \), where \( Q = 2^{b-1} - 1 \), using:
\begin{equation}
    z_{q,i} = \operatorname{clamp} \left( \operatorname{round}\left( \frac{z_i}{s_d} \right),\ -Q,\ Q \right).
\end{equation}

We store only the quantized vector \( \mathbf{z}_q = [z_{q,1}, z_{q,2}, \ldots, z_{q,n}] \in \mathbb{Z}^n \) and the corresponding scale factor \( s_d \), which significantly reduces storage compared to full-precision \texttt{float32} representations.

However, during training, neural networks such as ResNet-18 require full-precision inputs~\cite{huang2023normalization}. Therefore, we perform \textbf{dequantization} before feeding samples into the model. For each quantized sample, we reconstruct an approximate real-valued tensor \( \tilde{d} \in \mathbb{R}^{H \times W \times C} \) via:
\begin{equation}
    \tilde{z}_i = s_d \cdot z_{q,i}.
\end{equation}

The dequantized vector \( \tilde{\mathbf{z}} = [\tilde{z}_1, \ldots, \tilde{z}_n] \) is reshaped back to \( \tilde{d} \in \mathbb{R}^{H \times W \times C} \), and passed through the standard normalization pipeline. Since this dequantization is a simple element-wise operation performed once during data loading, it incurs negligible computational overhead.
 \begin{figure}[htbp]
    \centering
    \begin{subfigure}[b]{0.32\textwidth}
        \centering
        \includegraphics[height=5cm]{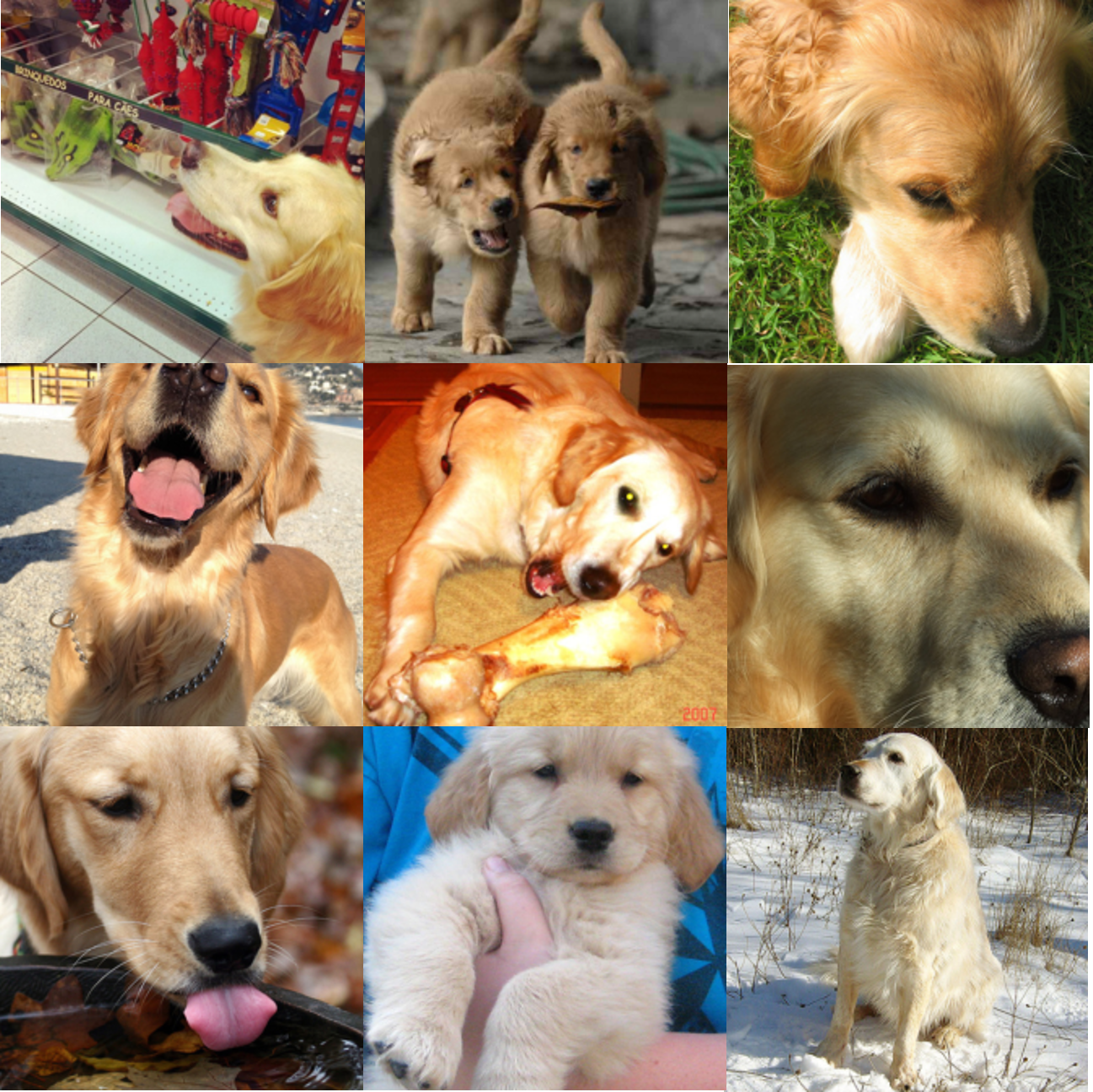}
        \caption{Original images with float32 dtype.}
    \end{subfigure}
    \hfill
    \begin{subfigure}[b]{0.32\textwidth}
        \centering
        \includegraphics[height=5cm]{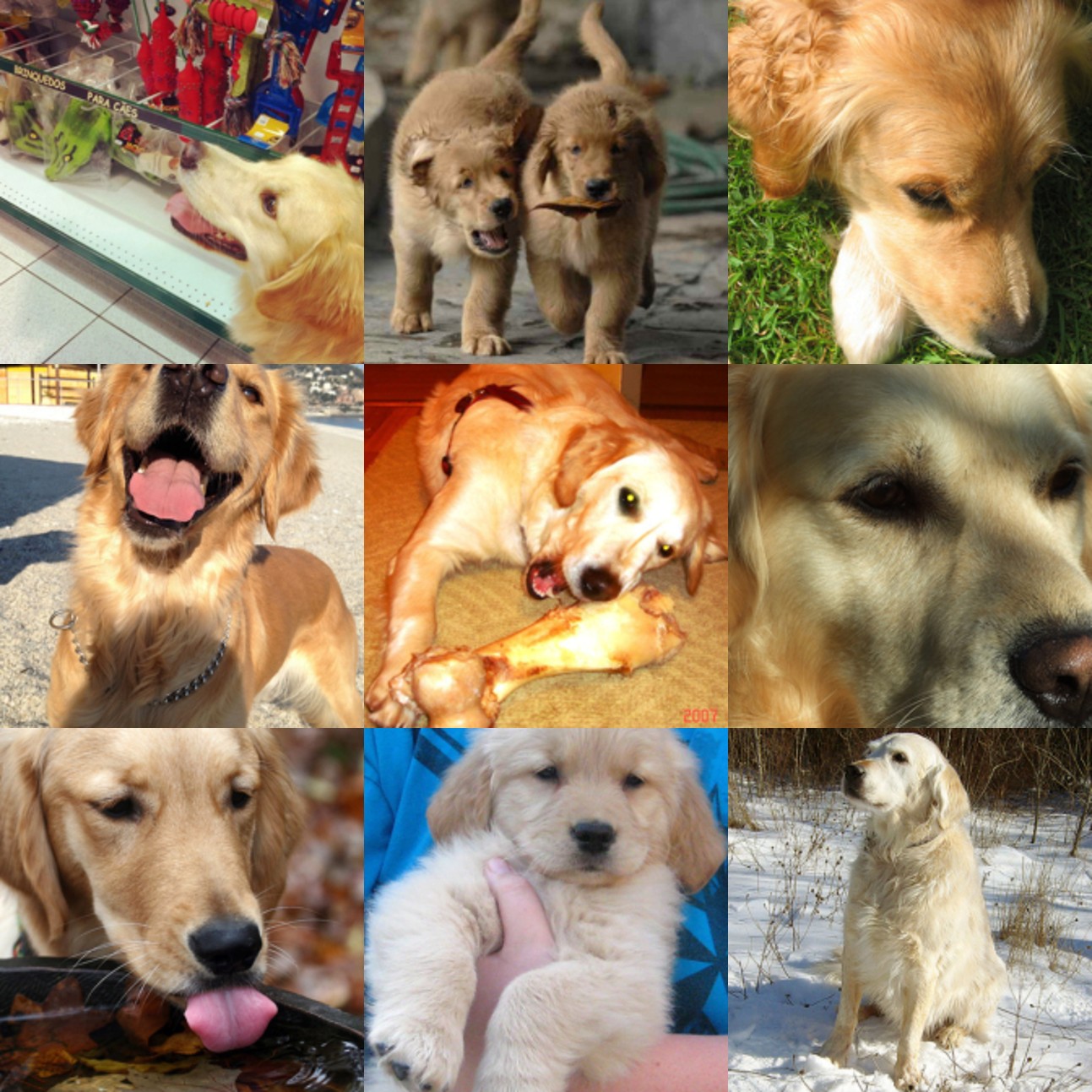}
        \caption{Images with 8-bit quantization and dequantization.}
    \end{subfigure}
    \hfill
    \begin{subfigure}[b]{0.32\textwidth}
        \centering
        \includegraphics[height=5cm]{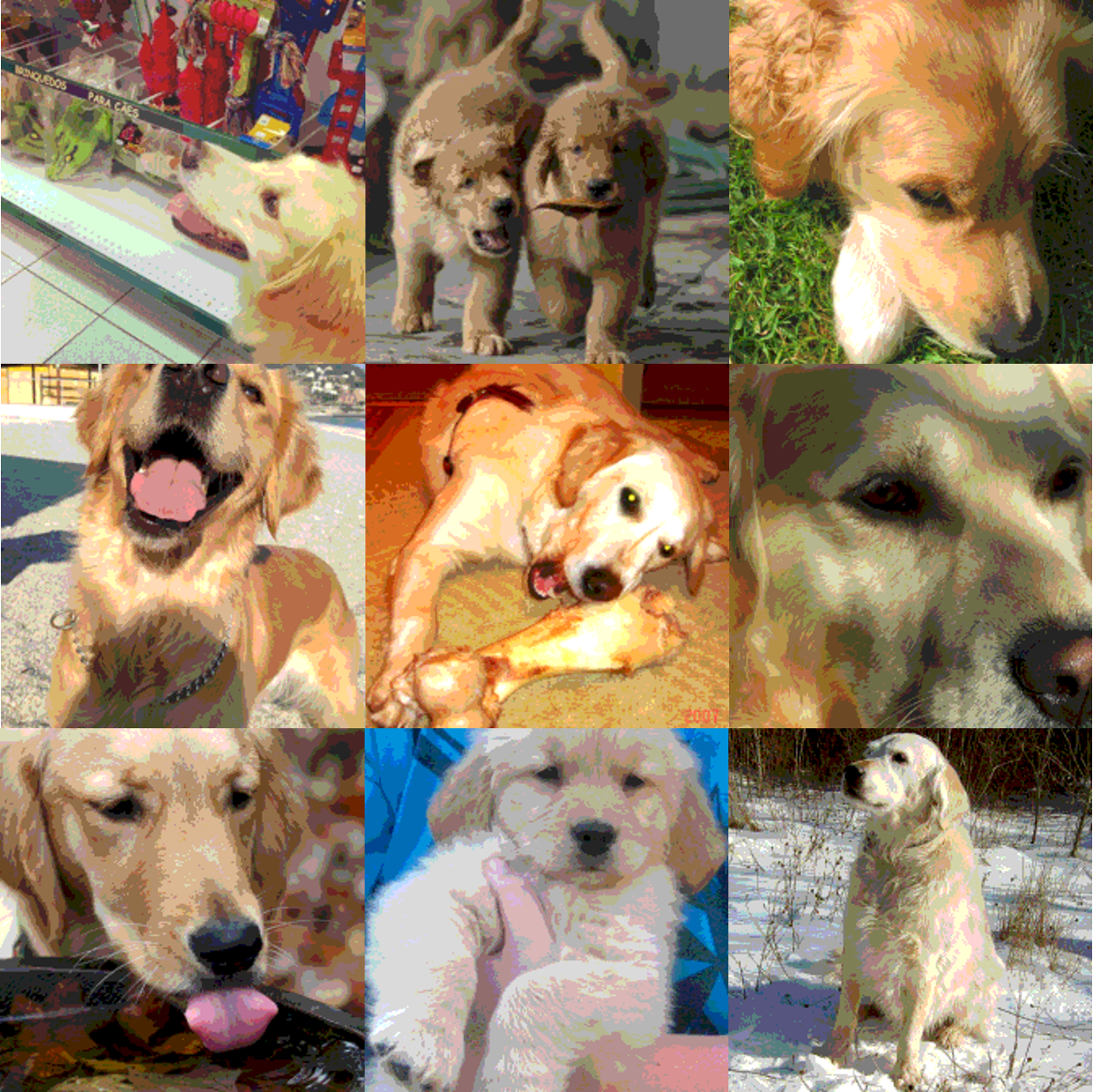}
        \caption{Images with 4-bit quantization and dequantization.}
    \end{subfigure}
    \caption{The left column displays the original images from the train-set. The center column shows the images reconstructed after 8-bit linear symmetric quantization, while the right column presents the results using 4-bit quantization. As observed, 8-bit quantization introduces negligible visual distortion and effectively preserves the perceptual quality of the original images, indicating that it is nearly lossless for this task. In contrast, 4-bit quantization introduces noticeable degradation; however, the primary visual structures and semantic content of the images remain largely intact and recognizable.}
\label{fig:int8/4}
\end{figure}
Fig. \ref{fig:int8/4} shows the visualization of the train-set after quantization and dequantization.
\subsection{Adaptive Quantization Bit-Width Allocation}

Uniformly applying a fixed quantization bit-width to all samples in a dataset is inefficient, as individual samples differ in their content and contribution to learning~\cite{wei2024advances}. Since quantization introduces \textit{lossy compression}~\cite{duan2023lossy}, it may degrade input feature quality and model performance, especially for samples sensitive to low-precision encoding.

Let \( \mathcal{D} = \{d_1, d_2, \ldots, d_N\} \) denote the training dataset, where each sample \( d \in \mathbb{R}^{3 \times H \times W} \) is a full-precision input tensor. For a neural network \( f(\cdot) \), we denote its extracted feature representation as \( f(d) \). After quantizing \( d \) with bit-width \( b \), we obtain its dequantized version \( \tilde{d} \), and the corresponding degraded feature \(  {f}(d; b) = f(\tilde{d}) \). The quantization-induced feature degradation is measured by:
\begin{equation}
\Delta_f(d; b) = \left\| f(d) - f(\tilde{d}) \right\|_2,
\end{equation}
where \( \|\cdot\|_2 \) denotes the Euclidean norm. Empirical results show that \( \Delta_f(d; b) \) varies across samples, indicating heterogeneous sensitivity to quantization. For example, on the CIFAR-10 training set with bit-width  \(b=4\), we observed that the feature degradation \( \Delta_f(d; 4) \)  spans a wide range, with the lowest 10\% of samples having values below 0.2, while the highest 10\% exceed 3.5. This large variance confirms that applying a uniform quantization bit-width is suboptimal. 

To capture this, we define a \textit{quantization sensitivity score} \( S(d) \) for each sample by measuring the gradient deviation between original and quantized inputs. Let \( \theta \) denote model parameters, \( \mathcal{L}(\cdot, y) \) be the loss function, and \( y \) the ground-truth label. The gradients with respect to original and quantized inputs are:
\begin{align}
g_{\text{orig}}(d) &= \nabla_\theta \mathcal{L}(f(d; \theta),\ y), \\
g_{\text{quant}}(d) &= \nabla_\theta \mathcal{L}(f(\tilde{d}; \theta),\ y),
\end{align}
where \( \tilde{d} \) is the dequantized sample after applying linear symmetric quantization to \( d \) under a predefined bit-width \( b \). The sensitivity score is then computed via cosine distance:
\begin{equation}
S(d) = 1 - \frac{\langle g_{\text{orig}}(d),\ g_{\text{quant}}(d) \rangle}{\|g_{\text{orig}}(d)\|_2 \cdot \|g_{\text{quant}}(d)\|_2}.
\end{equation}
A higher \( S(d) \) indicates stronger sensitivity to quantization, suggesting that low-bit compression may alter training dynamics for this sample.

We sort all samples in descending order of \( S(d) \), and split the dataset into two disjoint subsets:
\begin{equation}
\mathcal{D}_{\text{high}} = \{d \in \mathcal{D} \mid \text{top-}k\ S(d)\}, \quad \mathcal{D}_{\text{low}} = \mathcal{D} \setminus \mathcal{D}_{\text{high}}.
\end{equation}
We then assign higher bit-width \( b_{\text{high}} \) to the sensitive subset \( \mathcal{D}_{\text{high}} \), and lower bit-width \( b_{\text{low}} \) to the insensitive subset \( \mathcal{D}_{\text{low}} \), subject to a global bit-budget constraint:
\begin{equation}
\frac{1}{|\mathcal{D}|} \left( |\mathcal{D}_{\text{high}}| \cdot b_{\text{high}} + |\mathcal{D}_{\text{low}}| \cdot b_{\text{low}} \right) = b_{\text{avg}}.
\end{equation}
Here, \( b_{\text{avg}} \) denotes the target average bit-width across the dataset. The corresponding overall compression ratio is:
\begin{equation}
R = 1 - \frac{b_{\text{avg}}}{32},
\end{equation}
assuming that each original sample uses 32-bit floating-point representation.

\begin{figure}[h]
    \centering
    \begin{subfigure}[b]{0.48\textwidth}
        \centering
        \includegraphics[width=\linewidth]{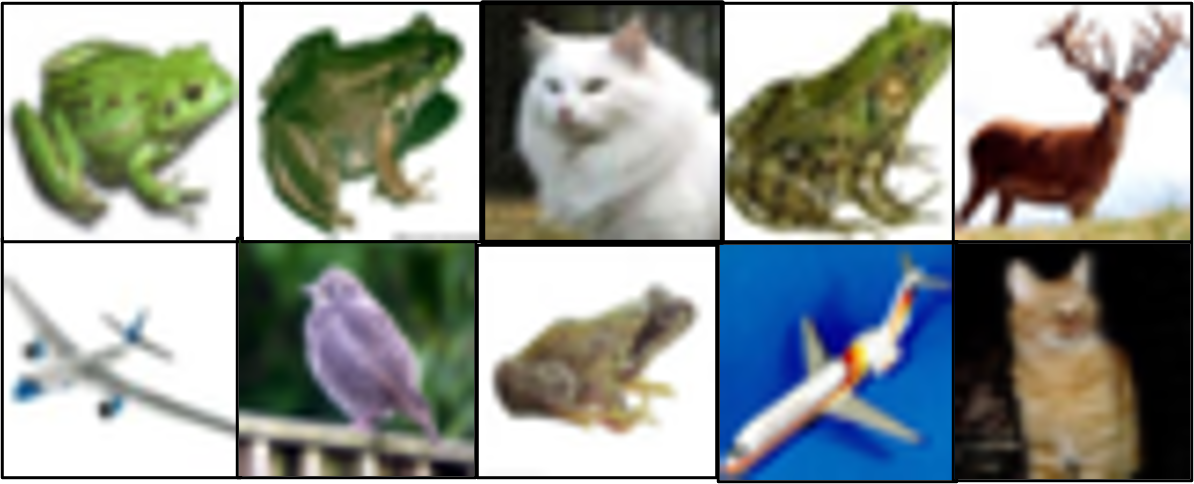}
        \caption{Images with smaller distance.}
        \label{fig:original}
    \end{subfigure}
    \hfill
    \begin{subfigure}[b]{0.48\textwidth}
        \centering
        \includegraphics[width=\linewidth]{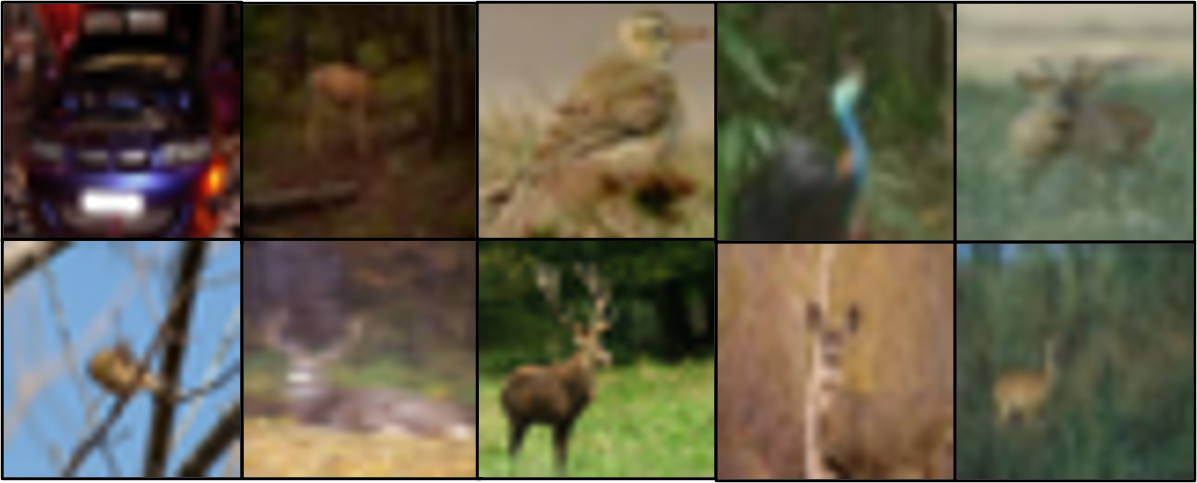}
        \caption{Images with larger distance.}
        \label{fig:quantized}
    \end{subfigure}
    \caption{Visualization of CIFAR-10 images with the top 10 smallest distances (a) and top 10 largest distances (b) under 8-bit quantization. Samples with small distances typically contain clear, large objects and simple backgrounds with prominent features. In contrast, samples with large distances often exhibit blurred objects and complex, textured backgrounds, making them more sensitive to feature loss from quantization. }
    \label{fig:comparison}
\end{figure}

\begin{figure}[h]
    \centering
    \includegraphics[width=0.8\textwidth]{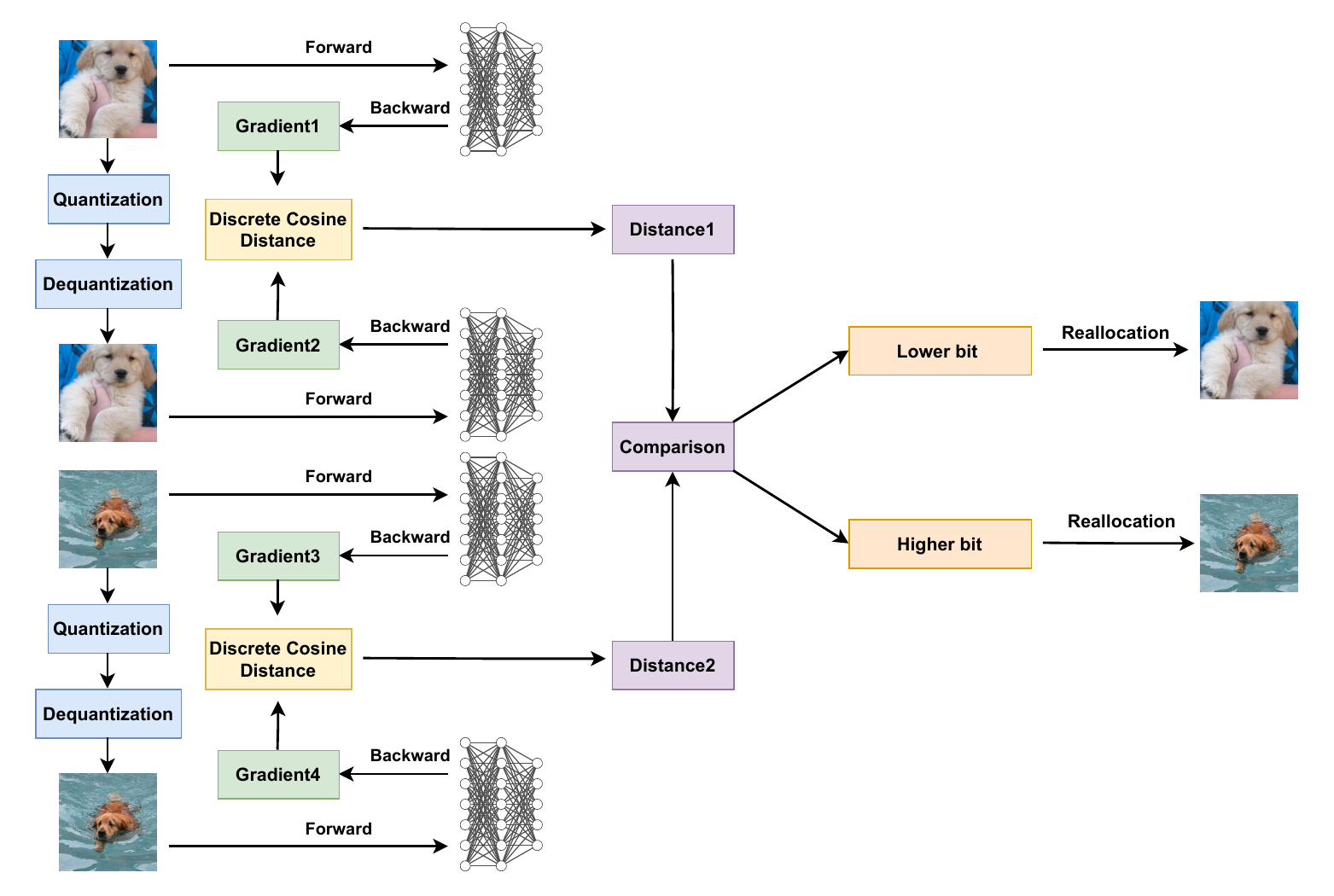}
    \caption{Overview of the Adaptive Quantization Framework
Each sample is initially quantized and dequantized using a uniform bit-width $b$. Based on the distance between the input gradient before and after quantization, bit-widths are then adaptively assigned—higher for samples with larger deviations, and lower for those with smaller deviations.}
    \label{fig:pdf_framework}
\end{figure}

We find empirically that setting \( |\mathcal{D}_{\text{high}}| = |\mathcal{D}_{\text{low}}| \) (i.e., equal split) achieves a good balance between compression and accuracy. This adaptive allocation strategy improves quantization efficiency by selectively preserving high-fidelity representations for crucial samples. Figure \ref{fig:comparison} shows the visualization of the samples with the largest distance and the smallest distance, and Figure \ref{fig:pdf_framework} shows the total framework of our adaptive quantization.

\begin{table}[ht]
    \centering
    \scriptsize
    \setlength{\tabcolsep}{4pt}
    \caption{Comparison of dataset pruning algorithms and our dataset color quantization algorithm on CIFAR-10, CIFAR-100, and ImageNet-1K with ResNet-18. CCS$_\mathrm{AUM}$, CCS$_\mathrm{Forg.}$, and CCS$_\mathrm{EL2N}$ refer to utilizing the AUM, Forgetting, and El2N scores and applying the CCS strategy for sampling. The model trained with the full CIFAR-10 and CIFAR-100 dataset achieves 95.45\% and 78.21\% accuracy, respectively. The model trained with the full ImageNet-1K dataset achieves 73.54\% accuracy.}
    \vspace{0.8em}
    \begin{tabular}{l|cccc|cccc|cccc}
        \toprule
            & \multicolumn{4}{c|}{CIFAR-10} & \multicolumn{4}{c|}{CIFAR-100} & \multicolumn{4}{c}{ImageNet-1K} \\
        Pruning Ratio    & 50\% & 75\% & 87.5\% & 93.75\% & 50\% & 75\% & 87.5\% &  93.75\%  & 50\% & 75\% & 87.5\% & 93.75\% \\
        \midrule
        Random &  93.40  & 89.37 & 80.15 & 77.04 &  71.07  & 61.72 & 45.19 & 39.71 &  70.34  & 63.28 & 53.34 & 50.32 \\
        Entropy &   92.11  & 81.88 & 67.58 & 68.51 &  68.26  & 51.38 & 39.77 & 28.96 &  70.76  &  58.32  & 42.04 & 31.04 \\
        EL2N &  94.80  & 80.68 & 35.19 & 21.31 &  68.10  & 31.16 & 20.36 & 8.36 &  67.17  & 40.87 & 15.99 & 11.69 \\
        AUM &  95.26  & 71.09 & 42.55 & 25.60 &  67.42  & 32.27 & 19.45 & 8.77 &  66.57  & 31.22 & 13.13 & 8.93 \\
        CCS$_\mathrm{AUM}$ & 94.93 & 91.81 & 87.61 & 75.77 & 74.45 & 65.05 & 52.16 & 30.02 & 70.52 & 65.44 & 59.96 & 45.58 \\
        CCS$_\mathrm{Forg.}$ & 94.03 & 90.05 & 87.66 & 75.31 & 74.38 & 66.45 & 51.86 & 31.12 & 69.97 & 65.04 & 58.57 & 45.28 \\
        CCS$_\mathrm{EL2N}$ & 95.04 & 91.07 & 87.05 & 75.09 & 74.79 & 65.75 & 52.16 & 32.07 & 68.95 & 64.45 & 57.94 & 45.19 \\
        TDDS (Strategy-E) & 95.50 & 92.72 & 87.44 & 77.97 & 74.07 & 67.44 & 57.13 & 52.15 &  68.69 & 66.02 & 57.95 & 45.99 \\
        $D^2$ & 94.91 & 92.54 & 87.88 & 78.55 & 75.91 & 68.49 & 58.01 & 52.79 & 71.82 &  66.83  & 58.01 & 46.78 \\
        \textbf{ADQ (Ours)} & \textbf{95.53} & \textbf{93.04} & \textbf{90.92} & \textbf{82.15} & \textbf{78.05} & \textbf{66.77} & \textbf{60.02} & \textbf{57.69} & \textbf{72.49} & \textbf{69.39} & \textbf{62.02} & \textbf{49.69} \\
        \midrule
        Full accuracy  & \multicolumn{4}{c|}{95.45\%}  & \multicolumn{4}{c|}{78.21\%} & \multicolumn{4}{c}{73.54\%}  \\
        \bottomrule
    \end{tabular}
    \label{tab:pruning_methods_combined}
\end{table}
\section{Experiments}
 
\subsection{Experiment Settings} 
 We evaluate the effectiveness of our proposed dataset color quantization method on three widely used image classification benchmarks: CIFAR-10, CIFAR-100~\cite{krizhevsky2009learning}, and ImageNet-1K~\cite{deng2009imagenet}, using the ResNet architecture~\cite{he2016deep} for all experiments. To demonstrate the efficiency of our approach, we compare it against several representative dataset pruning baselines, including EL2N~\cite{paul2021deep}, Entropy~\cite{coleman2019selection}, Forgetting~\cite{toneva2018empirical}, CCS~\cite{zheng2022coverage}, TDDS~\cite{zhang2024spanning} and $D^2$~~\cite{maharana2023d2}. These methods select a subset (coreset) from the training data to train the model and evaluate performance on the original test-set. For a fair comparison, all methods are evaluated under the same overall compression ratio. Given that the original data type is \texttt{float32}, we define the compression ratio \( R \) as \( R = 1 - \frac{b}{32} \), where \( b \) is the quantization bit-width. In our adaptive quantization setting (ADQ), different bit-widths are assigned to samples based on their sensitivity to quantization. Specifically, we evaluate the following \((b_{\text{high}}, b_{\text{low}})\) configurations: (16, 16), (10, 6),(8, 0), and (4, 0), where 0-bit indicates that the sample is dropped entirely corresponding to compression ratios of 50\%, 75\%, 87.5\%, and 93.75\%, respectively.
\subsection{Primary Results}

Table \ref{tab:pruning_methods_cifar_combined} reports the performance of various dataset pruning methods compared to our adaptive dataset quantization (ADQ) algorithm. ADQ consistently achieves the highest accuracy across all pruning ratios and datasets. At the extreme 93.75\% pruning ratio, ADQ reaches 82.15\% on CIFAR-10 and 57.69\% on CIFAR-100, outperforming strong baselines such as  $D^2$ (78.55\%, 52.79\%) and TDDS (77.97\%, 52.15\%). On ImageNet-1K, ADQ attains 49.69\% at 93.75\% pruning—exceeding CCS$_\mathrm{EL2N}$ (45.19\%) and $D^2$ (46.78\%). Even under a moderate 75\% pruning ratio, ADQ achieves 69.39\%, outperforming Entropy (58.32\%) and $D^2$ (66.83\%) by large margins. These results confirm that ADQ preserves more informative samples and delivers stronger generalization under aggressive compression.
Interestingly, we observe that under a 50\% compression ratio, the accuracy of the quantized model (95.53\%) slightly surpasses that of the full-precision model (95.45\%). This can be attributed to the implicit regularization and denoising effects introduced by quantization and dequantization. Such effects can improve generalization by removing redundant or noisy information from the input. 
\subsection{Additional Analysis}
\begin{table}[h]
    \centering
    \scriptsize
    \setlength{\tabcolsep}{0.4em}
    \caption{
    Combining color quantization and dataset pruning for an extreme compression ratio. The pruning rate is 90\%, and $b_q\%$ is the compression ratio using adaptive dataset quantization.
    }
    \vspace{0.8em}
    \begin{tabular}{l|cccc|cccc}
        \toprule
        & \multicolumn{4}{c|}{ResNet18, CIFAR-10} & \multicolumn{4}{c}{ResNet18, CIFAR-100} \\
        ACQ Compression ratio & $b_q=50\%$ & $b_q=75\%$ & $b_q=87.5\%$ & $b_q=93.75\%$ & $b_q=50\%$ & $b_q=75\%$ & $b_q=87.5\%$ & $b_q=93.75\%$ \\
        Total Ratio & 95.0\% & 97.5\% & 98.75\% & 99.3\% & 95.0\% & 97.5\% & 98.75\% & 99.3\% \\
        \midrule
        Random & 57.38 & 52.73 & 48.44 & 30.19 & 29.78 & 26.09 & 21.43 & 14.22 \\
        EL2N & 19.11 & 18.36 & 16.69 & 12.11 & 9.01 & 8.94 & 7.38 & 3.33 \\
        CCS$_\mathrm{AUM}$ & 66.95 & 62.44 & 52.10 & 44.50 & 25.64 & 25.08 & 24.28 & 18.36 \\
        TDDS & 65.78 & 57.68 & 49.42 & 41.55 & 24.91 & 23.77 & 23.12 & 16.51 \\
        \textbf{Ours + CCS$_\mathrm{AUM}$} & \textbf{82.92} & \textbf{82.53} & \textbf{79.01} & \textbf{68.73} & \textbf{49.26} & \textbf{43.22} & \textbf{42.49} & \textbf{28.94} \\
        \bottomrule
    \end{tabular}
    \label{tab:pruning_methods_cifar_combined}
\end{table}
\textbf{Combine with Dataset Pruning.} Since our approach is orthogonal to dataset pruning, we can combine adaptive dataset quantization with dataset pruning to achieve further compression.
Table \ref{tab:pruning_methods_cifar_combined} demonstrates the performance of combining two approaches to achieve extremely high pruning ratios such as 95.0\%, 97.5\%, 98.75\%, and 99.3\%.
Specifically, we integrate CCS-based coreset selection (10\% retention) with our color quantization, where image coresets are selected via CCS~\cite{zheng2022coverage} followed by $b_q\%$ compression ratio using adaptive dataset quantization. Table~\ref{tab:pruning_methods_cifar_combined} shows the performance of our method under extreme compression ratios, combining 90\% dataset pruning with adaptive color quantization. Across all settings, our approach (Ours + CCS$_\mathrm{AUM}$) significantly outperforms state-of-the-art pruning baselines. At the highest compression level (99.3\%), we achieve 68.73\% accuracy on CIFAR-10 and 28.94\% on CIFAR-100, outperforming CCS$_\mathrm{AUM}$ by over 24\% and 10\%, respectively. These results demonstrate the superiority of our method in highly constrained training scenarios.

\textbf{Comparison with Dataset Distillation.}
For  dataset distillation, we use G-VBSM~\cite{shao2024generalized}, SRe2L~\cite{shao2024generalized}, MTT~\cite{cazenavette2022dataset}, and DM~\cite{cazenavette2022dataset} as baselines. To match the compression rates in dataset distillation papers, we apply CCS$_\mathrm{AUM}$ with a pruning ratio \(r\%\) and use ADQ to compress the dataset with compression ratio $b_q$. For CIFAR-10 (IPC = 10, 99.8\% compression; IPC = 50, 99\%), we set \((r\%, b_q)\) to (99\%, 87.5\%) and (92\%, 87.5\%). For CIFAR-100 (IPC =10, 98\% compression; IPC=50, 90\% compression), we set \((r\%, b_q)\) to (84\%, 87.5\%) and (80\%, 50\%); For ImageNet-1K (IPC = 50, 96.1\%; IPC = 100, 92.1\%), we set \((r\%, b_q)\) to (70\%, 87.5\%) and (69\%, 75\%). This ensures consistency with the original paper while maintaining comparable compression settings. Table \ref{tab:Dataset_Distillation} demonstrates that our algorithm outperforms others.

\begin{table}[h]
\centering
\scriptsize
\caption{Comparison of dataset pruning algorithms and our algorithm on CIFAR-10 with ResNet-34 and ResNet-50.}
\vspace{0.3em}
\begin{tabular}{l|ccccc|ccccc}
\toprule
\multirow{2}{*}{Method} & \multicolumn{5}{c|}{\textbf{ResNet-34, CIFAR-10}} & \multicolumn{5}{c}{\textbf{ResNet-50, CIFAR-10}} \\
\cmidrule(lr){2-6} \cmidrule(lr){7-11}
& 80\% & 83\% & 87.5\% & 92\% & 96\% & 80\% & 83\% & 87.5\% & 92\% & 96\% \\
\midrule
Random       & 86.13 & 83.48 & 76.19 & 74.44 & 65.08 & 87.03 & 83.18 & 79.19 & 75.44 & 68.08 \\
EL2N         & 70.21 & 66.36 & 24.69 & 23.11 & 22.19 & 69.71 & 65.96 & 24.59 & 23.31 & 21.89 \\
AUM          & 58.34 & 49.59 & 29.92 & 25.03 & 21.37 & 57.04 & 49.29 & 29.09 & 25.60 & 21.35 \\
CCS$_\mathrm{AUM}$ & 89.34 & 87.92 & 87.52 & 74.01 & 41.22 & 89.44 & 88.11 & 86.75 & 74.31 & 71.02 \\
TDDS         & 90.58 & 88.15 & 86.92 & 73.05 & 40.98 & 91.10 & 89.02 & 86.92 & 73.22 & 69.00 \\
\textbf{Ours}         & \textbf{94.39} & \textbf{93.14} & \textbf{91.15} & \textbf{89.47} & \textbf{79.87} & \textbf{94.19} & \textbf{92.94} & \textbf{90.55} & \textbf{88.27} & \textbf{77.26} \\
\bottomrule
\end{tabular}
\label{tab:resnet34_resnet50_joint}
\end{table}
\begin{table}[h]
    \centering
    \normalsize
    \setlength{\tabcolsep}{0.3em}
    \caption{Comparison of dataset distillation algorithms and our algorithm on different datasets using ResNet-18. Entries with OOM denote Out of Memory.}
    \vspace{0.8em}
    \begin{tabular}{l|cc|cc|cc}
        \toprule
        & \multicolumn{2}{c|}{CIFAR-10} & \multicolumn{2}{c|}{CIFAR-100} & \multicolumn{2}{c}{ImageNet-1K} \\
        & IPC=10 & IPC=50 & IPC=10 & IPC=50 & IPC=100 & IPC=50 \\
        \midrule
        DM & OOM  & OOM  & 29.7 & 43.6 & OOM & OOM \\
        MTT & OOM  & OOM  & 40.1  & 47.7 & OOM & OOM \\
        SRe2L & 29.3  & 45.0  & 23.5 & 51.4  & 52.8  & 46.8 \\
        G-VBSM & 53.5 & 59.2  & 59.5 & 65.0  & 55.7  & 51.8  \\
        \textbf{Ours} & \textbf{56.7} & \textbf{61.3} & \textbf{60.4} & \textbf{66.8} & \textbf{58.8} & \textbf{55.3} \\
        \bottomrule
    \end{tabular}
    \label{tab:Dataset_Distillation}
\end{table}
\textbf{Network Generalization.} Table \ref{tab:resnet34_resnet50_joint} shows that our algorithm outperforms other algorithms on both larger networks like ResNet-34 and ResNet-50, highlighting its strong transferability across diverse neural network architectures.

\begin{table}[t]
    \centering
    \caption{Different ablation experiments about group strategies.}
    \vspace{0.8em}
    \begin{subtable}[b]{0.48\textwidth}
        \centering
        \setlength{\tabcolsep}{2pt}
        \caption{Different number of groups on CIFAR-10 on ResNet18.}
        \begin{tabular}{c c c c c}
            \toprule
            Compression Ratio & 50\% & 75\% & 87.5\% & 93.75\%  \\
            \midrule
            10 groups & 93.87 & 91.66 & 88.32 & 79.54 \\
            6 groups & 94.71 & 92.86 & 88.45 & 80.67 \\
            4 groups & 95.01 & 92.06 & 88.52 & 81.57 \\
            2 groups(ours) & \textbf{95.53} & \textbf{93.04} & \textbf{90.92} & \textbf{82.15}  \\
            \bottomrule
        \end{tabular}
        \label{tab:num}
    \end{subtable}
    \hfill
    \begin{subtable}[b]{0.48\textwidth}
        \centering
        \setlength{\tabcolsep}{2pt}
        \caption{Accuracy under fixed and adaptive ratio on CIFAR-100.}
        \begin{tabular}{c c c c c }
            \toprule
            Compression Ratio & 50\% & 75\% & 87.5\% & 93.75\% \\
            \midrule
            DBSCAN & 74.63  & 62.77  & 56.31  & 53.67 \\
            GMM & 74.78  & 62.65  & 56.42  & 53.89  \\ 
            k-means & 75.25  & 63.22  & 57.15 & 54.35   \\
            ours & \textbf{78.05} & \textbf{66.77} & \textbf{60.02} & \textbf{57.69} \\
            \bottomrule
        \end{tabular}
        \label{tab:cluster}
    \end{subtable}
    \label{tab:comfa}
\end{table}

\begin{table}[h]
    \centering
    \caption{Comparison of fixed and adaptive ratio allocation on CIFAR-10 and CIFAR-100.}
    \vspace{0.8em}
    \begin{subtable}[b]{0.48\textwidth}
        \centering
        \setlength{\tabcolsep}{3pt}
        \caption{Accuracy under fixed and adaptive ratio on CIFAR-10.}
        \begin{tabular}{c c c c c}
            \toprule
            Compression Ratio & 50\% & 75\% & 87.5\% & 93.75\%  \\
            \midrule
            Fix allocation & 95.09 & 91.66 & 86.51 & 78.37 \\
            Adaptive allocation & \textbf{95.53} & \textbf{93.04} & \textbf{90.92} & \textbf{82.15}  \\
            \bottomrule
        \end{tabular}
    \end{subtable}
    \hfill
    \begin{subtable}[b]{0.48\textwidth}
        \centering
        \setlength{\tabcolsep}{3pt}
        \caption{Accuracy under fixed and adaptive ratio on CIFAR-100.}
        \begin{tabular}{c c c c c }
            \toprule
            Compression Ratio & 50\% & 75\% & 87.5\% & 93.75\% \\
            \midrule
            Fix allocation & 73.22  & 64.17  & 57.85 & 50.85  \\
            Adaptive allocation & \textbf{78.05} & \textbf{66.77} & \textbf{60.02} & \textbf{57.69} \\
            \bottomrule
        \end{tabular}
         
    \end{subtable}
    \label{tab:comfa}
\end{table}

\subsection{Ablation Studies}
\textbf{Why use adaptive allocation?} Table \ref{tab:comfa} shows the comparison between using the adaptive allocation strategy; the result shows that the strategy can increase the performance of dataset compression significantly.

\textbf{Why choose to divide into two equal groups?} Table~\ref{tab:num} presents the results of using different numbers of groups during the adaptive quantization ratio allocation step. We observe that dividing the dataset into two groups yields the best performance. Table~\ref{tab:cluster} compares various strategies for grouping the dataset, including K-means~\cite{kanungo2000analysis}, Gaussian Mixture Models (GMM)~\cite{reynolds2015gaussian}, and DBSCAN~\cite{kulkarni2024survey}. The experimental results demonstrate that our proposed grouping strategy—based on comparing gradient differences before and after quantization—outperforms all conventional clustering methods.

\section{Conclusion}

Our work introduces a novel approach to dataset condensation by leveraging data quantization as a strategy to reduce dataset storage. This enhances data efficiency and opens new possibilities for optimized data representation.

Despite its effectiveness, our approach has several limitations and opportunities for future research. First, we currently employ linear symmetric quantization with a fixed quantization range. This can be improved by making the quantization ratio learnable for each image, potentially boosting dataset quantization performance. Second, our current method focuses solely on dataset quantization. In the future, we aim to develop a unified framework that integrates both dataset and model quantization to further improve the efficiency of AI systems. Third, we plan to extend our method beyond natural image datasets to other domains, such as medical imaging and synthetic datasets.

\appendix
\section{Experiments Setup}
 
\label{appendix:dataset}

\textbf{CIFAR-10/CIFAR-100} are datasets that consist of 10 and 100 classes, respectively, with image resolutions of $32 \times 32$ pixels.
We use a ResNet-18 architecture~\cite{he2016deep} to train the models for 40,000 iterations with a batch size of 256, equivalent to approximately 200 epochs. Optimization is performed using SGD with a momentum of 0.9, a weight decay of 0.0002, and an initial learning rate of 0.1. A cosine annealing scheduler~\cite{loshchilov2016sgdr} is applied with a minimum learning rate of 0.0001. Data augmentation includes a 4-pixel padding crop and random horizontal flips. The experimental setup follows that of~\cite{zheng2022coverage}.

\textbf{ImageNet-1K} is a dataset containing 1,000 classes and 1,281,167 images in total. The images are resized to a resolution of $224 \times 224$ pixels for training.
ResNet-34~\cite{he2016deep} is adopted as the network architecture. Models are trained for 300,000 iterations with a batch size of 256, corresponding to approximately 60 epochs. Optimization uses SGD with a momentum of 0.9, a weight decay of 0.0001, and an initial learning rate of 0.1. A cosine annealing learning rate scheduler is employed. The experimental setup follows that of~\cite{zheng2022coverage}.

\bibliographystyle{splncs04}
\bibliography{iccv_Security}
\end{document}